\pdfoutput=1
\documentclass[11pt]{article}

\usepackage{EMNLP2022}

\usepackage{times}
\usepackage{latexsym}
\usepackage[T1]{fontenc}
\usepackage[utf8]{inputenc}
\usepackage{microtype}
\usepackage{inconsolata}
\usepackage{url}
\usepackage{graphicx}
\usepackage{amsmath}
\usepackage{amsfonts}
\usepackage{tabularx}
\usepackage{booktabs}

\title{Multilingual News Location Detection using an Entity-Based Siamese Network with Semi-Supervised Contrastive Learning and Knowledge Base}

\author{
  Víctor Suárez-Paniagua \and Steven Derby \and Tri Kurniawan Wijaya \\
  Huawei Ireland Research Center \\
  Georges Court, Townsend St, \\
  Dublin 2, D02 R156, Ireland \\
  \texttt{\{victor.suarez.paniagua,steven.derby\}@huawei-partners.com} \\
  \texttt{tri.kurniawan.wijaya@huawei.com} \\
}

\begin{document}
\maketitle
\begin{abstract}
%
Early detection of relevant locations in a piece of news is especially important in extreme events such as environmental disasters, war conflicts, disease outbreaks, or political turmoils.
Additionally, this detection also helps recommender systems to promote relevant news based on user locations. 
%
In this paper, we present a language-agnostic system to detect relevant locations in a piece of news. 
Note that, when the relevant locations are not mentioned explicitly in the text, state-of-the-art methods typically fail to recognize them because these methods rely on syntactic recognition. In contrast, by incorporating a knowledge base and connecting entities with their locations, our system successfully infers the relevant locations even when they are not mentioned explicitly in the text.
%
To evaluate the effectiveness of our approach, and due to the lack of datasets in this area, 
we also contribute to the research community with a gold-standard multilingual news-location dataset, 
\textit{NewsLOC}.
It contains the annotation of the relevant locations (and their \emph{WikiData} IDs) of 600+ \emph{Wikinews} articles in five different languages: English, French, German, Italian, and Spanish. 
Through experimental evaluations, we show that our proposed system outperforms the baselines and the fine-tuned version of the model using semi-supervised data that increases the classification rate. The source code and the \textit{NewsLOC} dataset are publicly available for being used by the research community at \url{https://github.com/vsuarezpaniagua/NewsLocation}.
\end{abstract}

\section{Introduction}
In recent years, there has been a considerable increase in the creation of unstructured text data such as news and social media content, which has resulted in a greater need for effective strategies that are capable of extracting concrete entity classes from documents. For instance, \emph{Named Entity Recognition} (NER) --- systems used for extracting information relating to real-world objects such as people, companies, and locations --- has become an important topic within the field of \emph{Natural Language Processing} (NLP). In particular, the ability to effectively extract relevant locations in a text has numerous applications, including news recommendations and social media crisis event detection.

However, current systems are severely limited and in many cases, insufficient for handling many examples in the real world. Common approaches rely on the assumption that the place associated with the document is revealed at the surface level: the location is explicitly recorded within the text, for example, within a prepositional phrase or subordinate clause. In practice, this scenario is rarely guaranteed since such knowledge is implicitly encoded within semantically associated concepts --- it would be superfluous to state such information and in many cases, violate the Gricean maxims of quantity. For instance, from the following news title "Eiffel Tower evacuated after telephone bomb threat", we can infer the whereabouts (\textsc{Paris}, \textsc{France}) based purely on the intrinsic knowledge contained in the real-world entity \textsc{Eiffel Tower}.

Another typical issue with current systems is the scope of their granularity. Indeed, current approaches either provide extremely fine-grained geolocational information with limited expedience (such as longitude and latitude coordinates) or bounded coursed-grained labels which are insufficient for certain downstream applications. A system that can elucidate geographic knowledge from unstructured text data at multiple levels of granularity would be invaluable for many natural language applications.

In this work, we propose a novel approach for alleviating the shortcomings that generally plague such systems. 
First, we utilize the implicit knowledge from a pre-trained sentence transformer network to learn a novel model that can perform location disambiguation by learning to align salient entities in the text to place markers. 
Second, we introduce a novel self-supervised training strategy that allows our model to learn free-form location categories at multiple levels of granularity using data from an open knowledge base, which contains human-curated real-world properties. 
Finally, we provide a novel annotated dataset, which we call \textit{NewsLOC}, for evaluating the ability of future systems to effectively identify locations at different levels of specificity which are represented in the tuple form (city, country). We provide extensive details about our data collection with relevant statistics and in-depth quantitative analysis across a number of pre-trained sentence embedding models.

\section{Related Work}
Our work belongs to a larger body of research called \textit{geographic information retrieval} which aimed at extracting spatial information from documents, literature, news and other textual data. The field dates back to the 1960s and has a number of applications in fields of interest such as journalism, government policy and hazard response \citep{wang_spatiotemporal_2015}, and social sciences \citep{won_ensemble_2018,ehrmann_overview_2020}.

Many of the classical systems focus on the most fundamental problem of \textit{geoparsing} \citep{halterman_mordecai_2017,karimzadeh_geotxt_2019}, that is, detection of general location-related terms in the text along with precise linking of geographical positions in the form of coordinates or entries in geographical names databases (known as \textit{gazetteers}). Such systems, are usually built on top of NER models, which identify location-related entities. A slightly less general form of the geoparsing procedure --- consisting of location resolution of precise, unambiguous addresses or names of places --- is called \textit{geocoding}. This has recently been tackled using Neural Network architectures analysing the context of a location term occurrence and either directly regressing precise longitude/latitude coordinates \citep{radford_regressing_2021}, or finding an index in a multi-level cell decomposition of Earth's surface \citep{kulkarni_spatial_2020}. 
In a similar direction, \citet{hu_natural_2019} improve the current gazetteers with higher granularity location data by using \textit{Craigslist} housing market data and clustering techniques.


Geoparsing is often combined with the application of Knowledge Bases and ontologies, which serves the purpose of retrieval of more complex semantic patterns. 
For instance, \citet{wang_spatiotemporal_2015} builds up an ontology of hazard-related concepts leading to a mechanism allowing for the identification of hazard event information through linguistic processing. 
In addition, identifying location entities in tweets posted during crisis events is vital in extracting actionable situational awareness information. 
In this topic, \citet{khanal_multi-task_2021} propose a multi-task learning model using four datasets of crisis events.


In the literature, few works have covered Geolocation in pieces of news. 
A geographical news classification called Newsmap \citep{watanabe_newsmap_2018} uses a semi-supervised machine learning classifier to extract the geographical information in news and outperforms simple keyword matching or other systems such as Open Calais\footnote{\url{	http://www.opencalais.com/}} and Geoparser.io\footnote{\url{https://geoparser.io/}}. 
\citet{gupta_mapping_2020} fine-tune BERT \citep{devlin-etal-2019-bert} using a token-labelled corpus of news to extract their precise coverage together with SpaCy \citep{spacy} linguistic features and geocoding APIs such as OpenStreetMap\footnote{\url{https://www.openstreetmap.org/}} and Google Places API\footnote{\url{https://developers.google.com/maps/documentation/places/web-service/overview}}.
%
%
Another example is the work by \citet{sasaki_simple_2020} that recognize the salient locations to predict Japanese prefectures from articles using a NER model and a simple dictionary extracted from a Knowledge Base.

Recently, \citet{botha_entity_2020} propose a multilingual Entity Linking model 
using a dual encoder that maps the mention and entity representations to a similarity score. 
To evaluate this model, the authors extract the \emph{WikiData} ID of entities from \emph{Wikinews} articles and create a large dataset called \emph{Mewsli-9} in 9 languages and obtain 
state-of-the-art performance with one model able to link entities in 104 languages. 
Next, \citet{ruder_xtreme-r_2021} extend the dataset to 11 languages, called  \emph{Mewsli-X}, as a multi-task benchmark to evaluate the generalization of cross-lingual representations in multilingual tasks.

\section{Dataset}

We begin by outlining how we collected and processed news data for our work. \emph{Wikinews}\footnote{\url{https://www.wikinews.org/}} is an open collaborative journalism project that contains publicly-accessible news. Each article contains structured text (title, text, sources and categories), with additional supplementary material such as images, audio files and external links to \emph{Wikipedia} entities. In particular, the latter quality makes them particularly useful for building information extraction systems; in our case, we use this valuable source of information to create an annotated dataset for location detection in news.

\subsection{Information Extraction}
We begin by downloading the latest dump of news articles (version 2022/05/01)\footnote{\url{https://dumps.wikimedia.org/{CC}wikinews/20220501/wikinews-20220501-pages-articles.xml.bz2} where \{CC\} is the two-letter country code from the available languages} for German (de), English (en), Spanish (es), French (fr), and Italian (it). After that, we downloaded and patched the tool \emph{Wikiextractor}\footnote{\url{https://github.com/attardi/wikiextractor}} to extract and clean the wiki text from the database dumps, in the same way, at the \citep{botha_entity_2020} work. From here, we ran the \emph{Wikinews} parser\footnote{\url{https://github.com/google-research/google-research/tree/master/dense_representations_for_entity_retrieval/mel}} to extract entity-linked data from the processed news.

\subsection{Location Classification}
In recent years, knowledge graphs have become common use as lexical resources for many downstream natural language tasks. Knowledge graphs contain richly detailed entries of real-world concepts with corresponding semantic properties and links to other concepts, documented by humans. As such, they are useful tools for describing the meaning of entities and their associations with certain types of information; for instance, as a means to elucidate the location referenced in a document. In this work, we linked each category of the news to its appropriate ID and checked whether they are defined as a location using the knowledge graph. Here, the knowledge graph in question is \emph{DBpedia Ontology} and \emph{WikiData}, which provides us with properties to recognize categories as locations such as "populationTotal", "PopulatedPlace", "Location", "Place", and "Settlement", and the \emph{Wikipedia} properties; "country" (P17) and "located in the administrative-territorial entity" (P131). As we shall see, this will provide a valuable source of data for learning zero-shot location disambiguation at multiple levels of granularity.

\subsection{The \textit{NewsLOC} gold-standard}
To effectively estimate the quality of any system we also require an evaluation dataset of gold-standard annotations. This test data should contain several levels of geographical information for news in multiple languages, for which there are generally limited options in the wider literature. To create our test set, we gather $150$ news samples randomly for each language and automatically translated the non-English news for brevity. 
Two annotators are employed to label the data based on a simple guideline in order to annotate the main locations of the news (defined as (city, country) pairs) in a small set. 
Guidelines rules were updated accordingly, checking the disagreements between annotations before the annotators labelled a larger set of examples. The final dataset was constructed by merging the large and smaller sets, with non-agreeing examples either adjusted or removed. Table~\ref{tab:dataset} shows for each \emph{Wikinews} dump in different languages the number of documents, mentions parsed with the \emph{Wikiextactor} tool, unique entity \emph{Wikidata} IDs, the locations extracted from the categories using \emph{DBpedia}, the documents that contain locations, and the number of documents used for the test set.


\begin{table*}[!ht]
\centering
\begin{tabular}{lcccccc}
\hline
\textbf{Language} & \textbf{Documents} & \textbf{Mentions} & \textbf{Unique} & \textbf{Locations} & \textbf{Documents} & \textbf{Test}\\
\textbf{} & \textbf{} & \textbf{} & \textbf{entity IDs} & \textbf{in categories} & \textbf{with locations} & \textbf{documents}\\
\hline
English & 12,917 & 105,356 & 40,444 & 25,526 & 11,738 (\textit{90.87}\%) & 119 \\
French & 14,488 & 109,154 & 28,592 & 28,210 & 13,061 (\textit{90.15}\%) & 113 \\
German & 14,083 & 87,230 & 24,111 & 13,498 & 11,076 (\textit{78.65}\%) & 137 \\
Italian & 8,632 & 66,530 & 21,768 & 12,624 & 7,733 (\textit{89.59}\%) & 118 \\
Spanish & 11,234 & 81,981 & 23,239 & 23,077 & 10,282 (\textit{91.53}\%) & 131 \\
\textbf{Total} & 61,354 & 450,251 & 100,558 & 102,935 & 53,890 (\textit{87.83}\%) & 618 \\ 
\hline
\end{tabular}
\caption{The statistics for the \emph{Wikinews} dumps in different languages using the \emph{Wikiextractor} tool and the locations extracted from categories with the knowledge base approach.}
\label{tab:dataset}
\end{table*}

\section{Method}
\begin{figure*}[!ht]
    \centering
    \includegraphics[width=\textwidth]{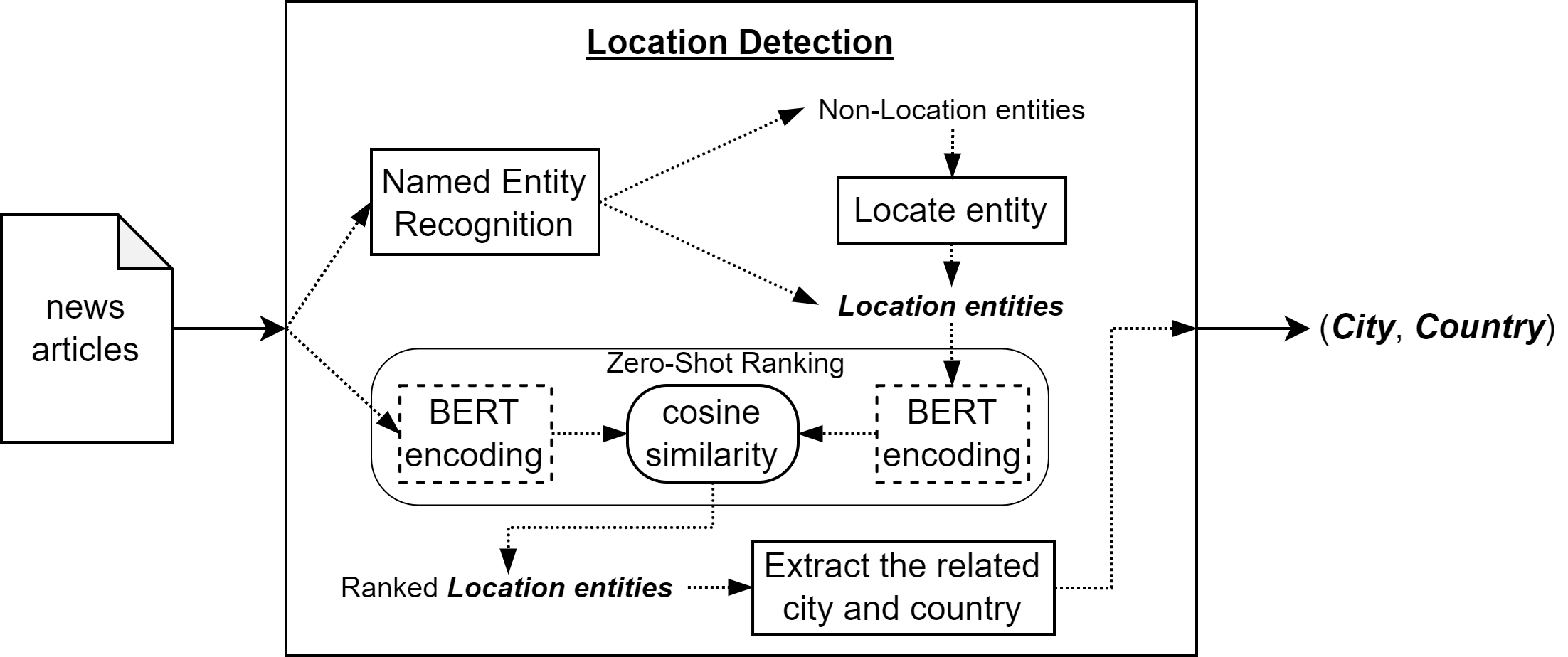}
    \caption{Overview of the Location Detection system}
    \label{fig:system}
\end{figure*}

The proposed system consists of two models: a NER model that extracts the mentions in news following previous approaches, and a Siamese Network that ranks the most relevant locations regarding their similarity with the article. Figure~\ref{fig:system} shows an overview of the process to detect the location of the news for any language.

\subsection{Named Entity Recognition in News}

Named Entity Recognition (NER) is a natural language processing task aimed at locating and classifying named entities in an unstructured text into a set of predefined classes such as person name, location, and organization. Such classification constitutes an important step in gathering structured information relevant to any document. As we have already suggested, named entities can be linked to some entities from a knowledge base which includes explicit location information relevant to our problem.

Since we wish to work with text in multiple languages, for NER in our model we use a state-of-the-art cross-lingual transformer \emph{XLM-Roberta Large} \citep{conneau_xlmr} trained on a corpus of 2.5TB of text in 100 languages and then fine-tuned for NER on standard datasets CoNLL3 \citep{conll2003} and OntoNotes 5.0 \citep{Weischedel2017OntoNotesA} -- and achieving approximately 92\% F1-score on those datasets. More precisely, our NER is a simple ensemble of the following two freely available models: 
\begin{itemize}
    \item XLM-Roberta Large CoNLL3 fine-tuned \footnote{\url{https://huggingface.co/xlm-roberta-large-finetuned-conll03-english}},
    \item TNER's \footnote{\url{https://github.com/asahi417/tner}} XLM-Roberta Large CoNLL3 and OntoNotes 5.0 fine-tuned \footnote{\url{https://huggingface.co/asahi417/tner-xlm-roberta-large-all-english}}.
\end{itemize}

In those two models, the location-related classes are given by \textbf{LOC} in CoNLL3, and \textbf{location} and \textbf{geopolitical area} in OntoNotes 5.0.

We insist on taking the ensemble to obtain higher coverage of entities in the processed text where the potential false positives are likely sieved out by the Contrastive Learning step.


\subsection{Implicit Knowledge Search}
Previous works have mainly focused on recognizing the location of the main text using only explicit location-based entities, although as we mentioned, such surface-level information may be absent from the text. Instead, we propose inferring the geographical location from non-location entities using information from the \emph{DBpedia} knowledge base. Concretely, the system looks for properties of the DBpedia page that contain "city", "country", "location", or "place" in their names, and extracts their values to locate the mention. For instance, \textsc{Queen Elizabeth II} will be situated in \textsc{Mayfair, London, United Kingdom} which is the "birthPlace" property of her \emph{DBpedia} page. Our approach allows us to determine the associated location from documents by searching the associated properties of entities in the text. 

Since we described a location as the (City/Town, Country) tuple, each entity in the article is related to its country and a city whenever possible. To do so, the system checks whether the \emph{WikiData} item of the location has the property "country" (P17), and inspects if the current location is a city, checking if any value in the property "instance of" (P31) contains the name "city", "capital", "municipality", "town", "village", or "commune", otherwise, the system recursively looks whether the following "located in the administrative territorial entity" (P131) is a city or not until the next item does not have this property.


\subsection{Contrastive Disambiguation Network}
While extracting second-order information from these entities can be effective for ascertaining salient geographical knowledge from the text, this information is both noisy and superfluous. Indeed, a document may contain numerous entities in the database while only a few of those may encode relevant information related to the location associated with the text. In order to effectively filter these conceptual features for only the most appropriate location properties, we propose a novel neural network-based architecture to learn the association between documents and salient locations within the document. 

Here, a Siamese Network creates a representation for each mention extracted by the NER model and, also, a representation for the whole text in the article. Then, the cosine function measures the similarity between each entity and the article representations. Thus, each mention in the text is ranked accordingly to its compatibility with the text.

\subsection{Entity Linking and Normalization}
For evaluation, once the locations are ranked, these entities need to be mapped to their common \emph{WikiData} ID in order to know if they are well categorized as an Entity Linking problem. For instance, the location \textsc{U.S.A.} should be the same as the location \textsc{the United States} which \emph{WikiData} ID is \textsc{Q30}. The ID associated with each extracted entity is taken from the \emph{WikiData} item information of its corresponding \emph{Wikipedia} page recovered using the entity as a query in the \emph{Wikipedia} search API\footnote{\url{https://{CC}.wikipedia.org/w/api.php?action=query&list=search&srlimit=max&srnamespace=0&format=json&srsearch={query}} where \{CC\} is the two-letter country code from the available languages and the \{query\} is the entity to look for}.

\begin{figure*}[!ht]
    \centering
    \includegraphics[width=0.825\textwidth]{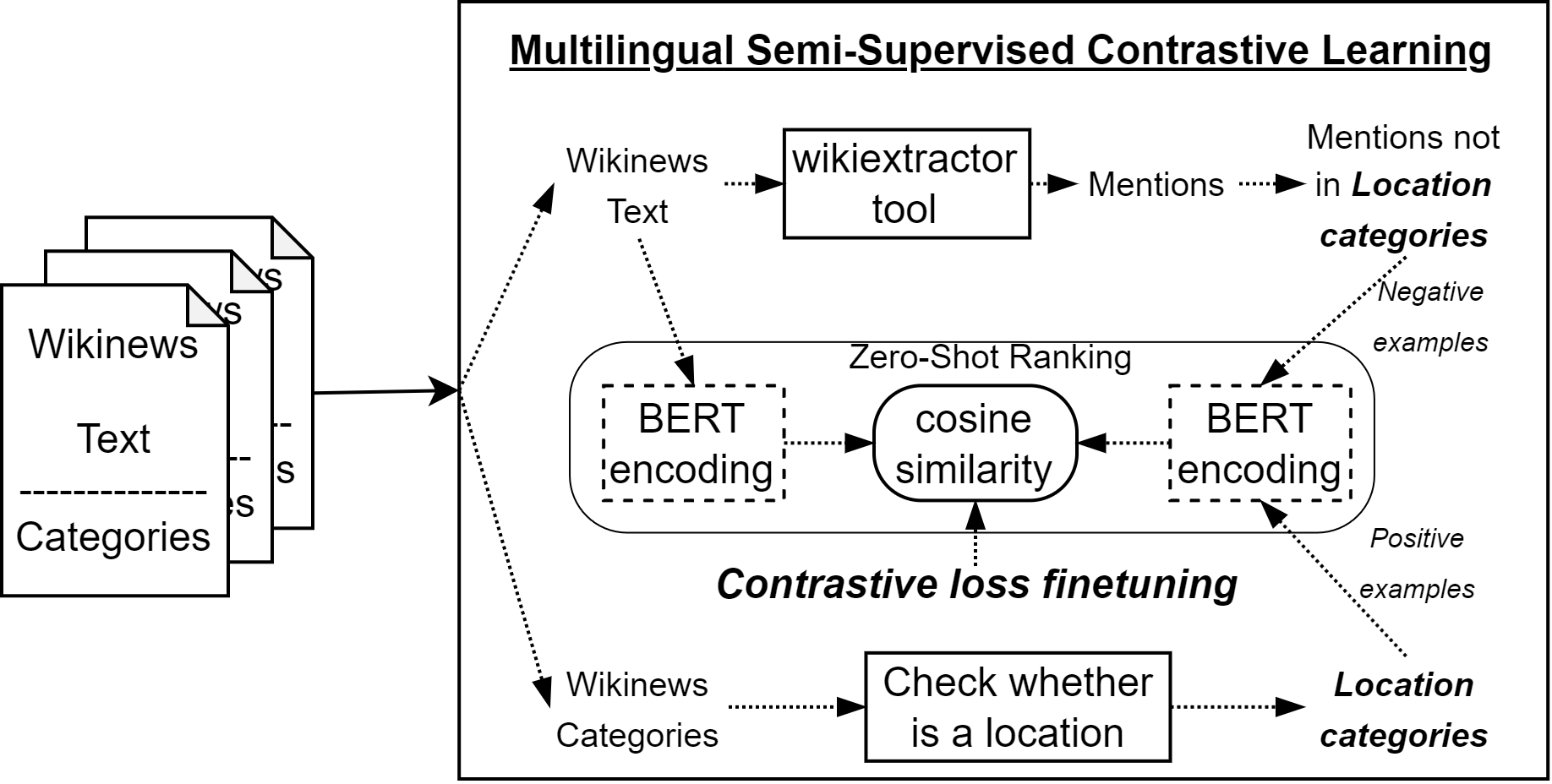}
    \caption{The Siamese Network Fine-tuning process}
    \label{fig:finetuning}
\end{figure*}

\section{Results and Discussion}
In this section, we analyze the proposed system for Location Detection and other modifications over the \emph{NewsLOC} test set. We measure the performance of the location detection using the \emph{Macro Precision@Top-1} (MP@1) in two levels: city-level and country-level, which evaluate whether the predicted city or country is presented in the gold standard location list, respectively.

\subsection{Baseline}
We follow two simple implementations as baselines: taking the first location or uncovered non-location that appears in the text as the main location. Even though the main information is contained at the beginning of the news, such as the title or the first paragraph, this is an inaccurate solution used to test the proposed method and its modifications. Table~\ref{tab:baseline} shows the performance of these two approaches. We observe that adding the located non-location does not increase the performance due to the fact that the first non-location mentions are not related to the most important locations. Besides, the country-level classification has better performance than the city-level which is a more fine-grained detection.

\begin{table}[!ht]
\centering
\begin{tabular}{lcc}
\hline
\textbf{Model} & \textbf{MP@1} & \textbf{MP@1} \\
\textbf{} & \textbf{country} & \textbf{city} \\
\hline
Baseline only locations & 67.54\% & 47.96\% \\
+ located non-locations & 67.41\% & 41.39\% \\
\hline
\end{tabular}
\caption{Performance in \emph{Macro Precision@Top-1} of the baseline models using only locations and adding the located non-location mentions for the country-level and city-level.}
\label{tab:baseline}
\end{table}

\subsection{Siamese Network}
The Siamese Network is implemented using the multilingual Sentence-BERT models \citep{reimers-2019-sentence-bert} which was fine-tuned as a teacher-student training method with mean pooling to aggregate feature vectors. Concretely, we measure the performance with the following models:

\begin{itemize}
    \item MPNet \citep{NEURIPS2020_c3a690be} and XLM-Roberta base \citep{conneau-etal-2020-unsupervised} fine-tuned using more than 50 languages\footnote{\url{https://huggingface.co/sentence-transformers/paraphrase-multilingual-mpnet-base-v2}}.
    \item MiniLM \citep{NEURIPS2020_3f5ee243} and Multilingual distilled MiniLM fine-tuned using more than 50 languages\footnote{\url{https://huggingface.co/sentence-transformers/paraphrase-multilingual-MiniLM-L12-v2}}.
    \item Multilingual Universal Sentence Encoder \citep{chidambaram-etal-2019-learning} and multilingual DistilBERT \citep{Sanh2019} fine-tuned with 15 languages\footnote{\url{https://huggingface.co/sentence-transformers/distiluse-base-multilingual-cased-v1}}.
    \item Same as previous model but fine-tuned with 50 languages\footnote{\url{https://huggingface.co/sentence-transformers/distiluse-base-multilingual-cased-v2}}.
\end{itemize}

Since many of the \emph{Wikinews} articles have more than the maximum sequence length tokens for these models, we implemented another solution instead of truncating the input of the article. 
For an article of tokens $S \in \mathbb{N}^{k}$ and Siamese Network $f\colon \mathbb{N}^{m} \rightarrow \mathbb{R}^{n}$, which can handle a max token length of $m$, the SpaCy multilingual sentence splitter \citep{spacy} divides the article $S$ into $p = \lceil k/m\rceil$ subgroups of sentences $\{S_{1}, S_{2}, \dots S_{p}\}$ such that $S = S_{1}S_{2}\dots S_{p}$ and $|S_{i}| <= m \, \forall 1 \leq i \leq p$. To retrieve the embedding representation $E$ of the article from the model, we average over all the subset representations as follows, 

\begin{equation}
    E = \dfrac{1}{p}\sum_{i=1}^{p}f(S_{i})
\end{equation}

Table~\ref{tab:siamese} presents the results in \emph{Macro Precision@Top-1} for the four models together with the modification of averaging all the sentence representations. As we can see, the best performance is obtained using the Sentence-BERT model Multilingual MPNet, which will be the base model for the following experiments. In addition, we can see that averaging the sentence representations does not overcome the models with truncation because the most informative part of the news is usually at the beginning of the article, such as the title and the first paragraph. In addition, the best of the four pre-trained models is Multilingual MPNet which is chosen to be the Siamese Network to be fine-tuned with the semi-supervised data.


\begin{table*}[!ht]
\centering
\begin{tabular}{l|cc|cc}
\toprule
\textbf{Model} & \textbf{MP@1} & \textbf{MP@1} & \textbf{MP@1} & \textbf{MP@1} \\
\textbf{} & \textbf{country} & \textbf{city} & \textbf{country} & \textbf{city} \\
\midrule \specialrule{.1em}{.05em}{.05em} 
& \multicolumn{2}{c}{Without Subdivisions} & \multicolumn{2}{c}{Average Subdivisions} \\
\midrule \specialrule{.1em}{.05em}{.05em} 
Multilingual MPNet & 68.2\% & 46.12\% & \textbf{68.46}\% & 45.99\% \\
Multilingual MiniLM & 67.28\% & \textbf{46.52}\% & 66.75\% & 45.99\% \\
Distiluse multilingual V1 & 68.33\% & 46.12\% & 68.07\% & 45.73\% \\
Distiluse multilingual V2 & 68.33\% & 46.12\% & 68.07\% & 45.99\% \\
\bottomrule
\end{tabular}
\caption{Performance in \emph{Macro Precision@Top-1} of the Sentence-BERT models and the aggregation of averaging the article subdivision representations for the country-level and city-level..}
\label{tab:siamese}
\end{table*}

\subsection{Knowledge Base}
The previous experiments only included the entities recognized as Location by the NER module. However, the remaining non-location entities may contain relevant information to detect the main location of the article. For this reason, we evaluate the performance of the Siamese Network comparison between the representations of the locations and the document, adding the non-locations, the located non-locations, the non-locations and the located non-locations joined by the string " in ", only the locations abstract and the locations and non-locations abstracts.

In Table~\ref{tab:kb}, the results using the six entity representations in the Siamese Network are listed for the four base models. The best performance is obtained using location and located non-locations that demonstrate that comparing the whole document against the entities using the Knowledge base information can efficiently represents the salience features of the mentions for Location Detection.

\begin{table}[!ht]
\centering
\begin{tabular}{lcc}
\hline
\textbf{Model} & \textbf{MP@1} & \textbf{MP@1} \\
\textbf{Multilingual MPNet} & \textbf{country} & \textbf{city} \\
\hline
only locations & 68.2\% & 46.12\% \\
non-locations & 64.91\% & 35.09\% \\
located non-locations & 70.57\% & 42.71\% \\
non-location "in" location & 66.1\% & 34.69\% \\
location \emph{DBpedia} abstracts & 69.12\% & 45.47\% \\
non-location abstracts & 65.44\% & 31.27\% \\
\hline
\end{tabular}
\caption{Performance in \emph{Macro Precision@Top-1} of different entity representation models using the Knowledge Base to locate the non-location mentions for the country-level and city-level.}
\label{tab:kb}
\end{table}

\subsection{Fine-tuned Siamese Network}
For the Siamese Network with Contrastive Learning, we propose to create semi-supervised data from pairs (document, location) labelled as positive using the categories that are classified as locations and pairs (document, entities) labelled as negative whether the entities in the document recognized by \emph{Wikiextractor} are not related to the extracted locations in the categories. We can see an outline of our training pipeline in Figure \ref{fig:finetuning}. By learning to associate entities that contain relevant location-based knowledge with the sentence --- aligning implicit knowledge from the network with real-world knowledge ---, we hope to ascertain a system that is capable of performing generalised inference, even for unseen locations. We trained the network during 32 epochs taking the best model using early stopping criteria over the validation set. We evaluate different Contrastive Learning loss functions to fine-tune the Siamese Network:
\begin{itemize}
    \item Cosine similarity loss: calculates the minimum square error between the $cosine$ similarity of the document $u$ and entity $v$ representations and the label of this pair $y$:
        \begin{equation}
        \mathcal{L}_{cos} = ||y - cosine(u,v)||_2
        \label{eq:cosine}
        \end{equation}
    
    \item Contrastive loss: reduces the $cosine$ similarity between the document $u$ and entity $v$ representations if the label is positive the, and increases the cosine similarity with a margin $m$ if the label is negative:
        \begin{equation}
        \mathcal{L}_{CL} = \left\{\begin{array}{rl}
\frac{cosine(u,v)^2}{2} & y=1 \\ \frac{max\{0,m-cosine(u,v)\}^2}{2}& y=0 \\
\end{array}\right.
        \label{eq:contrastive}
        \end{equation}
    \item Triplet loss: minimizes the euclidean distance between the document $u$ and a positive entity $v^+$ while maximizes the euclidean distance between the document $u$ and a negative entity $v^-$, and add a margin $m$:
        \begin{equation}
        \mathcal{L}_{Triplet} = max\{0,||u-v^+||-||u-v^-||+m\}
        \label{eq:triplet}
        \end{equation}
    \item InfoNCE loss: takes a batch of positive document-entity pairs, considers the remaining  pairs as negative, calculates the $cosine$ similarity between the document $u$ and each positive $v^+$ and negative $v^-$ entities, and minimizes the cross entropy loss:
        \begin{equation}
        \mathcal{L}_{NCE} = -log\frac{exp(cosine(u,v^+))}{\sum_i exp(cosine(u,v^i))}
        \label{eq:nce}
        \end{equation}
\end{itemize}

Since the performance usually increases with higher batch sizes for Contrastive Learning \citep{DBLP:journals/corr/HendersonASSLGK17}, we variate the number of samples used for each training step. Table~\ref{tab:finetune} presents the performance of the proposed model with different batch sizes. It can be observed that for all the Contrastive Learning losses increasing the batch number improves the performance, while the results are very similar for other losses. The experiments suggest that InfoNCE loss is the best loss function using the semi-supervised dataset from multilingual \emph{Wikinews} for Location Detection in news.

\begin{table*}[!ht]
\centering
\begin{tabular}{l|cc|cc|cc}
\toprule
\textbf{Model} & \textbf{MP@1} & \textbf{MP@1} & \textbf{MP@1} & \textbf{MP@1} & \textbf{MP@1} & \textbf{MP@1} \\
\textbf{Multilingual MPNet} & \textbf{country} & \textbf{city} & \textbf{country} & \textbf{city} & \textbf{country} & \textbf{city} \\
\midrule \specialrule{.1em}{.05em}{.05em} 
\textbf{Batch Size}& \multicolumn{2}{c|}{64} & \multicolumn{2}{c|}{128} & \multicolumn{2}{c}{256} \\
\midrule \specialrule{.1em}{.05em}{.05em} 
Cosine similarity loss & 66.75\% & 35.35\% & 68.2\% & 35.09\% & 67.94\% & 35.48\% \\
Contrastive loss  & 68.59\% & 37.98\% & 69.51\% & 38.63\% & 68.46\% & 37.06\% \\
Triplet loss  & 65.44\% & 35.48\% & 66.62\% & 34.95\% & 67.15\% & 35.48\% \\
InfoNCE loss & 71.48\% & 44.81\% & \textbf{71.75} \% & \textbf{46.91}\% & 71.35 \% & 45.47\% \\
\bottomrule
\end{tabular}
\caption{Performance in Macro Precision@Top-1 of different Contrastive Learning loss function with different batch sizes for the country-level and city-level.}
\label{tab:finetune}
\end{table*}


\section{Conclusion and Future Work}
As far as we know, an Entity-Based Siamese Network never has been used to detect the main location of a news article in multiple languages. Moreover, the \emph{DBpedia} and \emph{WikiData} Knowledge bases are valuable language-agnostic sources of information to locate all the entities recognized as non-locations. Furthermore, training this model with semi-supervised Contrastive Learning increases the recognition of the current Location of the news. It demonstrates much stronger performance over the baseline given that it can predict the most relevant Location from the context of the text. To test our novel approach, we also propose a new dataset called \textit{NewsLOC} where 618 \emph{Wikinews} articles are classified into different Locations as (city, country) tuples. The performance of the proposed system is tested and analyzed using this multilingual dataset showing promising results for Location Detection in news.
For future work, we would like to increase the \textit{NewsLOC} dataset for non-European languages. Since the Semi-supervised dataset work very well, we want to expand the languages to all the available languages in Wikinews to create a fully multilingual Location Detection system. In addition, we saw that the NER is a very important component in the proposed system, thus, we propose to create an end-to-end system that not only extracts locations but also ranks them updating the representation from scratch.

\section*{Limitations}
Since the proposed system utilizes the knowledge base as the component to extract the information for locations and to locate the entities that are not locations, we rely upon the following tools that could affect the performance of the present work:
\begin{itemize}
    \item \emph{Wikinews} is a \emph{Wikimedia} project that can be edited by users and could contain some conflicts. However, all the news has one or more sources to verify the information presented in the article and the \emph{Wikinews} policies include ways to open discussion between authors about these special cases.
    \item Some articles contain the editor'of s choice location, this means that at the very beginning of the news the city and country may appear together with the date of the article. As this is included in the text the system also recognizes them as an entity and can be part of the ranked locations.
    \item \emph{Wikiextractor} is a parser generated with rules, but not all the articles in \emph{Wikinews} follow the same templates. In fact, this tool seems to be optimized for the English version and we found some parsing errors in other languages.
    \item The system extracts the title for \emph{DBpedia} and the \emph{Wikidata} ID of the entities querying the \emph{Wikipedia} search engine tool. In some cases, the first page recovered may not be exactly the same one as expected.
\end{itemize}

\section*{Ethics Statement}
All articles were collected from Wikinews which is a publicly available source of news. The information about the entities is extracted from DBpedia and Wikidata which are also publicly available. In this sense, we do not use any confidential personal information. In order to reduce the possible bias in the articles, we use different languages and also applied a random selection from all the collections of news to avoid malicious or discriminatory articles in our dataset. To be completely transparent about the dataset annotated, we share the Wikinews URLs of the articles and the labels to be used freely by the research community.

\section*{Acknowledgements}
The authors would like to thank the members of the Artificial Intelligence Research Group in the Huawei Ireland Research Center for their valuable discussion and comments. Especially to the AINews group, and to Maciej Zdanowicz and Long Mai who collaborated in the annotation process of the presented dataset.

\bibliography{Location_Detection}
\bibliographystyle{acl_natbib}

\appendix

\end{document}